\documentclass[conference]{IEEEtran}
\IEEEoverridecommandlockouts

\usepackage{cite}
\usepackage{amsmath,amssymb,amsfonts}
\usepackage{algorithmic}
\usepackage{graphicx}
\usepackage{textcomp}
\usepackage{makecell}
\usepackage{xcolor}
\usepackage{hyperref}
\def\BibTeX{{\rm B\kern-.05em{\sc i\kern-.025em b}\kern-.08em
    T\kern-.1667em\lower.7ex\hbox{E}\kern-.125emX}}
\begin{document}

\title{Disentangling Representation using Attributes-based Gaussian Estimation for Medical Sound Diagnosis

}


\author{\IEEEauthorblockN{Ke Zhao}
\IEEEauthorblockA{\textit{Shenyang Institute of Computing Technology, Chinese Academy of Science}, Shenyang 110168, China\\
\textit{University of Chinese Academy of Science}, Beijing 100049, China \\
zhaoke22@mails.ucas.ac.cn}

}

\maketitle

\begin{abstract}
Deep learning has a powerful capability of feature extraction. However, the lack of fairness and interpretability in deep neural networks poses limitations to their adoption in the medical domain. This paper proposes a disentangled representation learning (DisenRL) framework, named the Attributes-based Gaussian Estimation for Disentangled Representation (AGEDR), which incorporates Attribute Mapping Embedding (AME) modules designed to map attributes into vectors and align them with a subset of the latent vectors in a Variational AutoEncoder (VAE). This part of the latent vector will be disentangled from the remaining latent vectors by minimizing mutual information. A classifier is then trained using the mean parameters of the latent vectors from the VAE. Extensive experiments demonstrate that AGEDR outperforms both conventional classification models and existing disentangled representation learning methods. The ablation experiments also indicate the disentangling capability and fairness of AGEDR. 
The source code is publicly available at \href{https://github.com/ZhaoKe1024/DisentangledRepr}{https://github.com/ZhaoKe1024/DisentangledRepr}
\end{abstract}


\section{Introduction}
\label{sec:intro}
The extensive application of machine learning and deep learning in medical signal analysis has significantly contributed to the development of medical diagnosis and medical equipment. Nevertheless, machine learning demands high-quality features, while deep learning lacks interpretability for automatically extracted features. Given the wide variations in human constitution, differences in age, gender, lifestyle, medical history, and regional environments may affect model performance, resulting in reduced applicability. This phenomenon is commonly referred to as domain shift and poses a major challenge for domain generalization. Current deep learning algorithms often exhibit these problems in practical applications, exhibiting biases towards individuals with different genders or medical histories, as well as lifestyle habits.

Disentangled representation learning (DisenRL) is a technique that uses specific methods to obtain latent vectors with decomposable structures, addressing the issues of the biases in the model and domain shift in the prediction task. Supervised disentangled methods, in particular, enable the alignment of latent vectors with actual semantics, thereby genuinely solving the problem of lack of interpretability.

To incorporate additional attribute features other than the data labels into the latent vectors, a common approach is to concatenate the attributes into the latent vectors as InfoGAN\cite{10.5555/3157096.3157340}. Wu et al. directly trained multiple prediction models to derive the latent vectors corresponding to different attributes\cite{wu2024explainable}. GSL sets an exchangeable structure for the representations and obtains latent vector with equivariance during training \cite{ge2021zeroshot}. Kota Dohi et al. employed a Normalizing Flow model, incorporating an additional rotational speed attribute, and combined them through information-theoretic methods to achieve parameter disentanglement \cite{Dohi2021DisentanglingPP}. It is evident that DisenRL inevitably necessitates the introduction of prior knowledge, including attributes or structures, which have been proven essential\cite{locatello2019challenging}. The above methods all reflect this idea, which disentangles the attributes by fitting the distribution of attributes so as to improve the generalization on the data of different attributes and solve the problem of domain shift between different attribute values.

After introducing additional attributes, it is necessary to align them with a part of the latent vector obtained by audio feature extraction, i.e., to disentangle them from the rest of the latent vector. InfoGAN disentangles the labels and latent vectors by minimizing mutual information (MI) using evidence lower bounds (ELBO), thereby obtaining an equivariant structure \cite{10.5555/3157096.3157340}. IDB-SR aiming to disentangle sensitive attributes, trains a sensitive attribute prediction task in conjunction with MI disentanglement \cite{Lu2023InformationBD}. Furthermore, to ensure orthogonality, i.e., disentanglement, between different attributes, diagonalization of the correlation coefficient matrix can also be adopted\cite{Lv2022CausalityIR}. From the perspective of data augmentation, constructing sample-label pairs with equivariant transformations to derive equivariant relationships, the disentangled representation models can be obtained\cite{qi2020learning,qi2019avt}.

For classification tasks, if the attributes of data samples are given, the information carried by that attribute is inherently present within the data samples. Through deep neural networks and classification task training, latent vectors related to that attribute can be extracted. However, these latent vectors are not subject to any prior distribution, making it impossible to perform probabilistic density estimation as in Variational AutoEncoder (VAE)\cite{kingma2014autoencoding}. This limitation hinders the calculation of relative entropy (i.e., Kullback-Leibler divergence ($D_{KL}$)) and mutual information (MI)\cite{45903}. This paper proposes a method that comprises two parts: the Attribute Mapping Embedding (AME) modules and a VAE. The AME is responsible for mapping scalar values of different categories into normal parameters, which are then used to obtain the attribute latent vector $\hat{z}^{\alpha}$ through the reparameterization trick. The VAE, on the other hand, extracts the latent vector $z$ consisting of two parts $\{z^{\beta},z^{\alpha}\}$ from the original data, which serve as the input of the VAE's Decoder. 
We refer to $z^{\alpha}$ as the attribute latent representation, and the non-attribute latent vector $z^{\beta}$ can be viewed as the joint distribution encompassing all attributes other than the $z^{\alpha}$. They are disentangled using MI minimizing. This framework is named Attributes-based Gaussian Estimation for Disentangled Representation (AGEDR).

The contributions of this paper are as follows:

\begin{enumerate}
    \item We propose an Attribute Mapping Embedding (AME) module that transforms scalar semantic attributes into Gaussian latent representations, to restore the deep features of attributes and simplify the calculation of $D_{KL}$, which is used to align the latent vector with the part of the latent vector of VAE.
    \item Based on the above, we introduce the concepts of attribute latent vectors and non-attribute latent vectors and disentangle them using the ELBO of mutual information minimizing. A series of experiments are conducted on the COVID-19 cough dataset COUGHVID to compare the proposed method with previous approaches.
    \item Through a series of comparative experiments and two ablation experiments, we not only illustrate the classification performance of AGEDR, but also provide empirical evidence of attribute-related information reduction, 
    so as to mitigate attribute-related domain shifts and reduce unnecessary sensitive information in the dataset.
\end{enumerate}

\section{Proposed Method}
\label{sec:pagestyle}
The method proposed in this paper is a model based on the Variational AutoEncoder (VAE) architecture, where the latent vectors follow a normal distribution $N(0, 1)$. To embed the semantics of specific attributes within the latent vectors, semantic alignment and disentangled representation are achieved through information-theoretic approaches.

\subsection{Attributes Mapping Embedding}
\begin{figure}[ht]
\center
\includegraphics[width=3in]{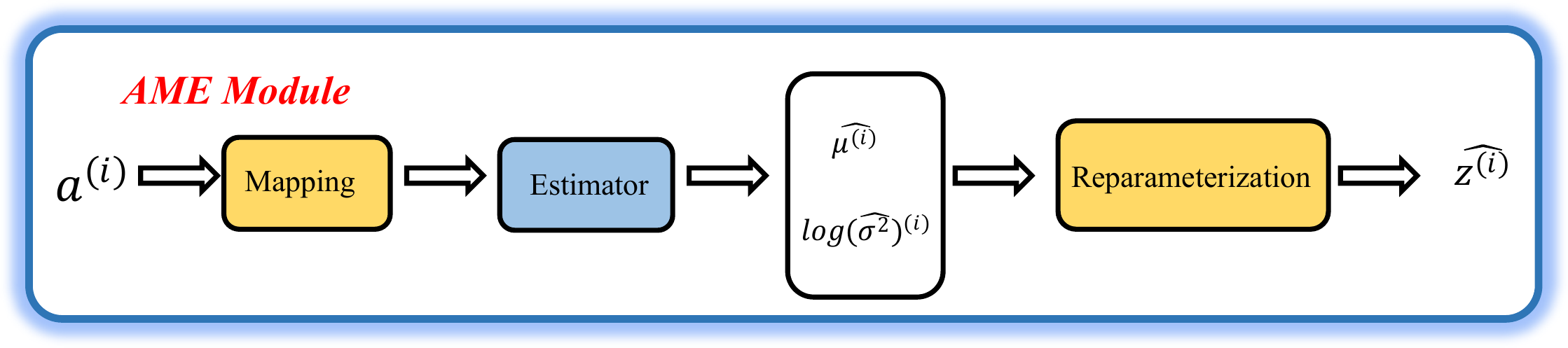}
\caption{ The AME module comprises three components: (1) the scalar-valued attributes are transformed into vectors through a mapping function as Eq.\ref{eq1}. (2) an estimator, which is a two-layer neural network to estimate the parameters $\hat{\mu^{(i)}}$ and $\hat{\sigma^{(i)}}$ of a normal distribution. (3) The latent vector $\hat{z^{(i)}}$ is derived through the reparameterization trick.}
\label{fig1:amemodule}
\end{figure}

Similar to data annotations, attributes are often presented in scalar type. To transform them into deep representations that contain richer information, a mapping function with learnable parameters is employed to convert scalar attribute values into vector types, then, the latent vector is obtained through a multi-layer neural network\cite{karras2019style}, which is used to adjust the density distribution of the vector to fit the Gaussian distribution. This module is called Attributes Mapping Embedding (AME), as shown in Fig.\ref{fig1:amemodule}. The mapping function is described as Eq.\ref{eq1}, and Eq.\ref{eq2} denotes the reparameterization trick:

\begin{equation}
\begin{aligned}
        (\hat{\mu^{(i)}}, \hat{\sigma^{(i)}})&=\left\{ \begin{array}{c}
	est(M^{dis}\cdot onehot(a^i)) \,\,\quad \quad, if\,\,a\in N_+\,\,\\
	est(M^{con}\cdot a^i) \quad \quad \quad \quad \quad \quad \quad  , if\,\,a\in R\\
    \end{array} \right., \\
    i &= 1,2,\dots , r,
    \label{eq1}
\end{aligned}
\end{equation}

\begin{equation}
    \begin{aligned}
        \hat{z^{(i)}} = \hat{\mu^{(i)}} + \hat{\sigma^{(i)}} \odot \epsilon^{(i)}, i=1,2,\dots ,r,  \\
        \epsilon^{(i)} \sim N(0,1),
    \end{aligned}
    \label{eq2}
\end{equation}

Where $M^{dis} \in \mathbf{R}^{n^{(i)} \times c^{(i)}}$ is a matrix, with $c^{(i)}$ representing the number of categories for the categorical attribute $a^{(i)}$, and $r$ being the number of attributes in the dataset. $n^{(i)}$ denotes the dimensionality of the resulting vector after transformation, i.e., $\hat{z^{(i)}} \in \mathbf{R}^{n^{(i)}}$. The function $onehot(\cdot)$ converts scalar values into one-hot vectors to select a column from the matrix $M^{dis}_{n^{(i)} \times c^{(i)}}$ as a vector. $M^{con}$ is a column matrix that transforms continuous attributes into a vector. The function $est(\cdot)$ represents a neural network used for estimating Gaussian parameters $\hat{\mu^{(i)}}$ and $\hat{\sigma^{(i)}}$, and subsequently, the latent vector $\hat{z^{(i)}}$ is derived through the reparameterization trick\cite{kingma2014autoencoding}, as the Eq.\ref{eq2}.

Assuming the dimensionality of the latent vector $z$ in VAE is $N$, we partition it into two components: the attribute latent vector $z^{\alpha}$ and the non-attributes latent vector $z^{\beta}$. The $z^{\alpha}$ encompasses $r$ latent vector $z^{(i)}, i=1,2,\dots,r$ with corresponding dimensions $n^{(1)}, n^{(2)}, \dots, n^{(r)}$ by concatenating them by dimension, as shown in Eq.\ref{eq6}. The non-attribute latent vector $z^{\beta}$ contains all the information in sample $X$ except the above r attributes, which is a part that lacks explicability, and its dimension is $n^{(0)}$. This gives us the equation as Eq.\ref{eq3}:

\begin{equation}
    \sum_{i=0}^{r}n^{(i)}=N,
    \label{eq3}
\end{equation}
\begin{equation}
    z^{\alpha} = concat(z^{(1)},z^{(2)},\dots,z^{(r)}),
    \label{eq6}
\end{equation}

\begin{figure*}[ht]
\center
\includegraphics[width=6in]{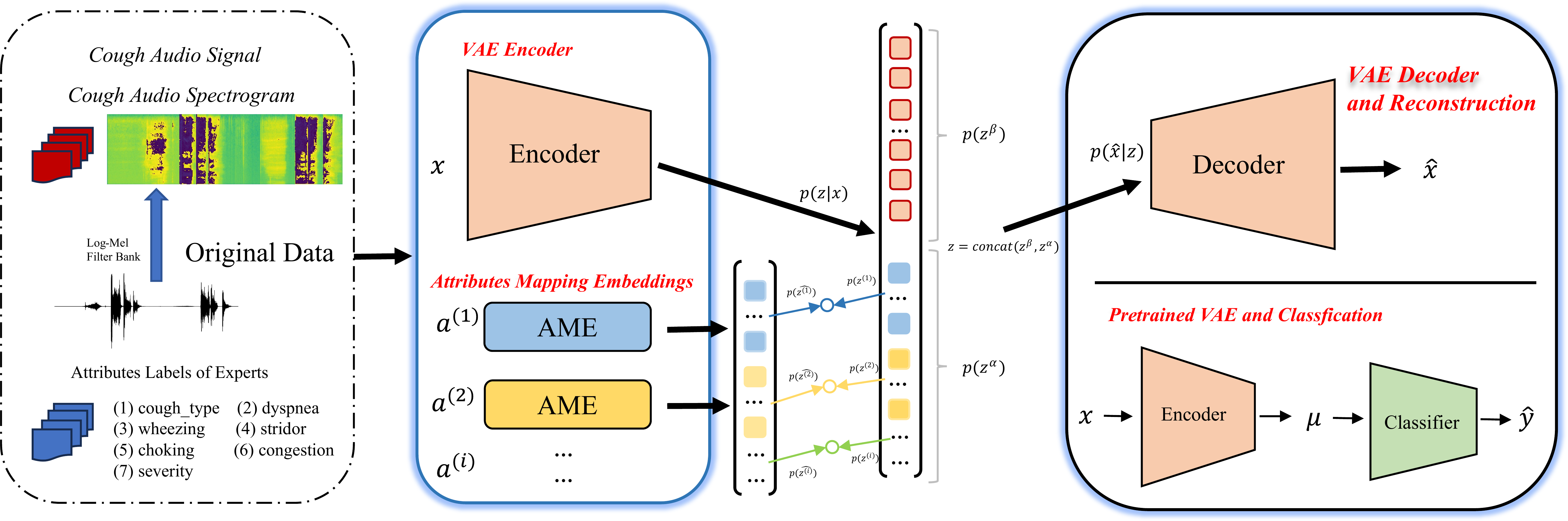}
\caption{ Proposed Framework: (1) The employed COUGHVID dataset\cite{orlandic_coughvid_2021} comprises audio signals and annotations for seven attributes. The output of the encoder is a latent vector $z$. Meanwhile, the attributes undergo transformation via the AME module into a vector $\hat{z}^\alpha$. Both latent vectors, $z^{\beta}$ and $z^\alpha$, are obtained through the reparameterization trick. Alignment between $\hat{z}^\alpha$ and $z^{\alpha}$ is achieved by minimizing the $D_{KL}$.(2) The VAE is trained through a reconstruction task.(3) After pre-training, the normal mean parameters output by the encoder are extracted and utilized as the input for the final disease prediction task.}
\label{fig2:framework}
\end{figure*}

In order for VAE to learn the distribution obtained by the AME module, the alignment of the two can be achieved directly by minimizing $D_{KL}$, so that VAE can directly extract all attribute information contained in $z^{\alpha}$ from the original data without additional attribute annotation. The $D_{KL}$ is proposed as the regular term of the loss function as Eq.\ref{eq4}:
\begin{equation}
    \begin{aligned}
        L_{attri}^{(i)} &= \sum_{i=1}^{r}D_{KL}(z^{(i)}, \hat{z^{(i)}}) \\
        &= \sum_{i=1}^{r}(log{\frac{\hat{\sigma^{(i)}}}{\sigma^{(i)}}}) + \sum_{i=1}^{r}(\frac{(\sigma^{(i)})^2+(\mu^{(i)}-\hat{\mu^{(i)}})^2}{2(\hat{\sigma^{(i)}})^2})-\frac{r}{2},
    \end{aligned}
    \label{eq4}
\end{equation}


\subsection{Disentangled Latent Vector}
\label{sec:typestyle}

Although we distinguish between attribute latent vector and non-attribute latent vector through the above definition, which respectively represent the vector segment containing some attributes and the other hybrid vector segments, as shown in formula \ref{eq5}, they are still entangled before trying to disentangle them, because the input samples $X$ encompass information about all attributes of the data, there exists information overlap between the latent vectors obtained from attribute encoding $z^{\alpha}$ and the remaining latent vector $z^{\beta}$. This overlap not only reduces the robustness of the model but also diminishes its interpretability. To mitigate this entanglement, mutual information ($MI$) minimization is employed to decrease the information overlap between the two \cite{10.5555/3157096.3157340, Lu2023InformationBD}, serving as the disentanglement loss function as Eq.\ref{eq7}.

\begin{equation}
    z = Encoder(X) = concat(z^{\beta}, z^{\alpha}),
    \label{eq5}
\end{equation}
\vspace{-0.5cm}
\begin{equation}
    \begin{aligned}
        MI(z^{\alpha}, z^{\beta}) = MI(z^{\alpha}, x) +MI(z^{\beta}, x)- MI(z,x),
    \end{aligned}
    \label{eq7}
\end{equation}

The MI is difficult to calculate due to the unknown joint distribution of $z^{\alpha}$ and $z^{\beta}$, and in our method, the length $n^{(i)}$ of the different attribute latent vector $z^{(i)}$ is different (depending on the number of categories of a certain attribute), so we can not directly calculate the KL divergence of the two as a regular term to control the entanglement degree of the two.
Because the mutual information between \(z_\alpha\) and \(z_\beta\) is difficult to compute directly, we optimize a evidence lower bound (ELBO)\cite{Lu2023InformationBD} following, a variational method of neural estimation for MI.
Therefore, the loss function Eq.\ref{eq7} is transformed into the following loss function $L_{disen}$ as Eq.\ref{eq8}.

\begin{equation}
    \begin{aligned}
    MI(z^{\alpha}, z^{\beta}) &\leq L_{Disen} \\
        &=  \mathbb{E}_{x\sim p(x)}E_{x' \sim p(x)}KL(p(z^{\alpha}|x)||p(z^{\alpha}|x')) \\
        &+ \mathbb{E}_{x\sim p(x)}E_{x'\sim p(x)}KL(p(z^{\beta}|x)||p(z^{\beta}|x')) \\
        &- \mathbb{E}_{x\sim p(x)}E_{z^{\alpha},z^{\beta}\sim p(z|x)}logq(x|z),
    \end{aligned}
    \label{eq8}
\end{equation}

where the first and second terms are derived based on the definition of mutual information combined with Jensen's inequality\cite{Lu2023InformationBD}, where p(x) represents the input sample set and $p(z^{\alpha}|x)$ signifies the process of extracting the latent vectors $z^{\alpha}$ from samples $x$ using an encoder, sample $x'$ represents another batch of samples, with similar interpretations for the latent vector $z^{\beta}$. The third term can be obtained through the ELBO of MI\cite{10.5555/3157096.3157340}, $q(x|z)$ is considered as the process of reconstructing $x$ from the latent vector $z$.


\subsection{The Classification Task}
Due to the presence of reparameterization, the latent vectors of VAE carry randomness caused by noise $\epsilon^{(i)}\sim N(0, 1)$, which is beneficial for generation tasks but has a negative impact on prediction tasks\cite{higgins2017betavae}. Therefore, for the trained VAE model, the mean parameter $\mu$ output by the encoder is directly utilized as the input for the classifier, and the classification loss function is denoted as $L_{cls}$, In this paper, we use the focalloss function\cite{lin2017focal} for the classification task. The framework of the final model is depicted in Figure \ref{fig2:framework}. Finally, we obtain the loss function as Eq.\ref{eq9} consisting of four terms, where $w_1, w_2, w_3, w_4, v_1, v_2, v_3$ represent the weights of different terms.

\begin{equation}
    \begin{aligned}
        L_{total} &= w_1 L_{cls}+w_2L_{vae} + w_3\sum_{i=1}^{r}L_{attri}^{(i)} - w_4L_{Disen}\\
        &= w_1 L_{cls}+w_2L_{vae} + w_3\sum_{i=1}^{r}L_{attri}^{(i)} \\
        &- w_4v_1MI(z^{\alpha}|x)- w_4v_2MI(z^{\beta},x) \\
        &+w_4v_3MI(z,x),
    \end{aligned}
    \label{eq9}
\end{equation}

\section{Experiments and Results}
\subsection{Dataset}
The dataset employed in this paper is COUGHVID\cite{orlandic_coughvid_2021}, which comprises 34,434 cough audios and the corresponding user's own annotation obtained through network crowdsourcing. In addition, there are more than 2800 samples with expert annotation. Since the data are not collected by professional means, there are many unreliable samples. Orlandic et al.\cite{orlandic_coughvid_2021} further annotated the data set and gave the probability that an audio contains cough samples. In this paper, the data is clear according to this field. The steps are as follows:
\begin{enumerate}
    \item Only the columns that retain all expert annotations are left.
    \item Filter out samples without field "status".
    \item Filter out samples whose "cough\_detected" field is less than 0.8.
    \item Filter out samples whose "quality" field is not "good".
\end{enumerate}

After filtering, 720 samples remained. Preprocess the 720 samples by following these steps. First, the audio sample is divided into multiple cough audio segments according to the mute clips, and the cough audio less than 0.36s is discarded. This procedure yielded 2,850 cough segments. Then, set the standard audio sample length to 32360 and the sampling rate to 22050, i.e., 1.46s. For longer ones, take the middle 1.46s and fill 0 on both sides of the insufficient length.

Finally, 2076 samples labeled "health" and 774 samples labeled "covid-19" were obtained. Then the audio signal is converted into a log-Mel spectrogram, and the size of the spectrum is 64 in length and 128 in dimension. In addition, select "cough\_type" and "severity", and the number of categories is 3 and 4, respectively.

\subsection{Model Settings}
\begin{figure}[ht]
\center
\includegraphics[width=3.5in]{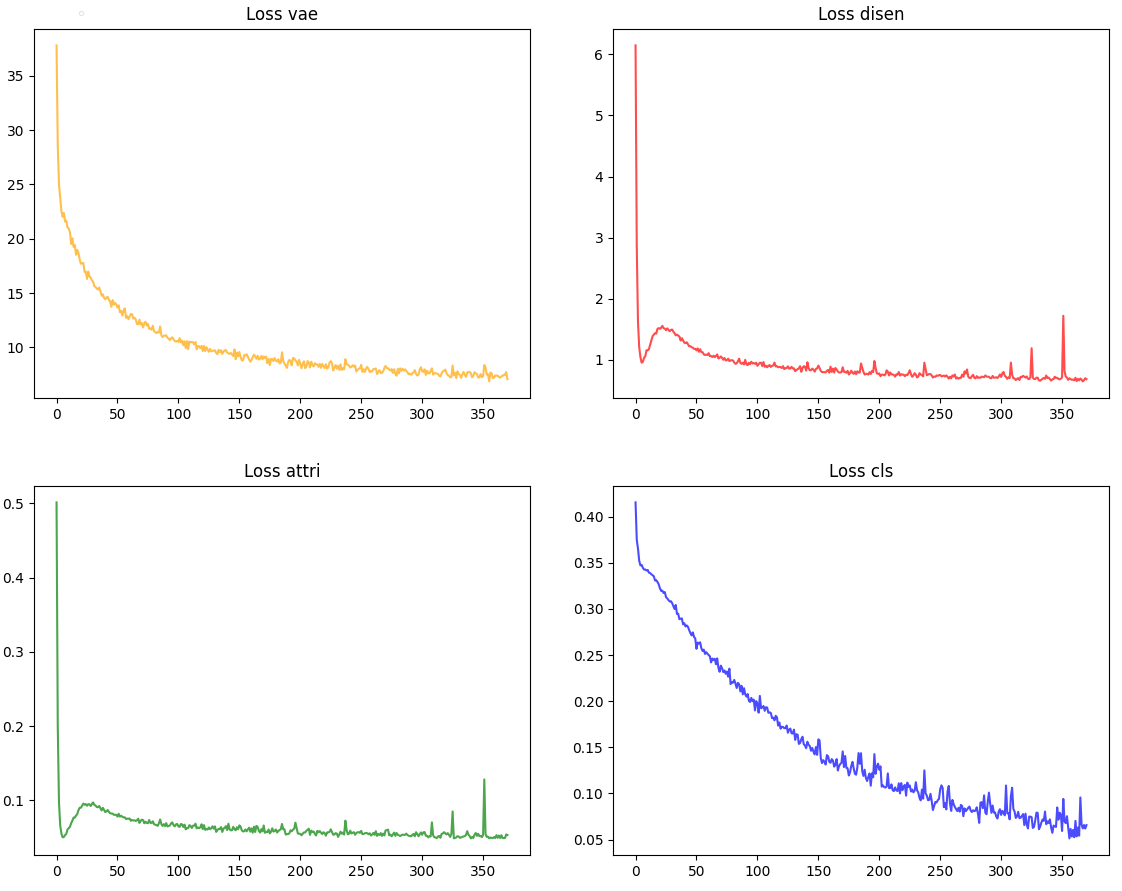}
\caption{The decreasing trend of each part of the loss function in the iteration process. (a) Is the loss function value $L_{vae}$ of VAE, (b) is the value of the loss function $L_{disen}$ part, (c) is the value of the loss function $L_{attri}$ part, (d) is the value of the classification loss $L_{cls}$.}
\label{fig3:lossdecrease}
\end{figure}

We employ convolutional networks to implement the VAE, where the latent vector is designed to have 30 dimensions. Among them, 6 dimensions correspond to the 'cough\_type' attribute, and 8 dimensions correspond to the 'severity' attribute, leaving 16 dimensions for non-attribute latent vectors. The classifier also adopts a two-layer fully connected neural network by default. The weights $w_1, w_2, w_3,w_4$ are set to 2.0, 0.3, 0.0025, and 0.01, and $v_1,v_2,v_3$ are set to 1.0, 1.25, and 5.0, respectively. The learning rates of the AME module, VAE, and classifier are 0.0003, 0.0001, and 0.0002, respectively. All modules are optimized using the Adam optimizer\cite{kingma2014adam}. Due to the imbalanced class distribution in the classification task, the ratio of health and COVID-19 samples is about 2.68:1. To mitigate this issue, the classification loss function $L_{cls}$ adopts the focal loss function\cite{lin2017focal}. After 370 epochs of training, the model parameters tend to converge, and the values of each part of the loss function are shown in Figure \ref{fig3:lossdecrease}.

\subsection{Performance Comparision}
For comparative performance evaluation on the classification task, this paper selects classification models MobileNetV2\cite{sandler2018mobilenetv2}, AST\cite{gong21b_interspeech}, and DisenRL-based methods DisenIB\cite{pan2021disentangled} and IDB-SR\cite{Lu2023InformationBD} as baselines.

The results of precision, recall, and AUC (Area under ROC) are presented in Table.\ref{tab1:comp}. It can be observed that AGEDR outperforms the classification models such as MobileNetV2 and AST. The group using latent vectors $z=concat(z^{\alpha}, z^{\beta})$ achieved an accuracy of 92.52\% and the highest recall of 99\%. Compared to similar DisenRL-based methods as DisenIB and IBD-SR, higher performance has been achieved. Further, the classification heads of AGEDR were tested using a 2-layer neural network and SVM, similar performance was achieved, SVM even achieved higher classification ability, indicating the effectiveness of various classifier types.

\begin{table}[ht]
  \caption{ Classification ability of different models \\ on the validation set.}
	\centering
	\begin{tabular}{p{22mm}|lll}
		\hline
		\makecell{\textbf{Models}} & \textbf{Precision} & \textbf{Recall} &\textbf{AUC}\\
        \hline
        \makecell{MobileNetV2}  & 88.42\% & 84.68\%  & 86.51\%  \\ 
        \hline
        \makecell{AST} & 91.32\%  & 82.24\%  & 86.54\%  \\ 
        \hline
        \makecell{DisenIB } & 90.12\%  & 91.58\%  &  90.84\% \\ 
        \hline
        \makecell{IDB-SR}  & 94.23\%  & 92.43\%  & 93.32\%  \\
        \hline
        \makecell{AGEDR \\ NN(\{$z^{\beta},z^{\alpha}$)\}} & 92.52\%  & \textbf{99.00\%}  & 95.50\%  \\ 
        \hline
        \makecell{AGEDR \\ SVM(\{$z^{\beta},z^{\alpha}$\})}  & \textbf{97.23\%}  & 96.65\%  & \textbf{97.06\%}  \\
        \hline
	\end{tabular}
	\label{tab1:comp}
\end{table}

\subsection{Disentanglement Study}
To evaluate the disentangled ability of the AME module, we conduct an experiment by ignoring the attribute latent vector $z^{\alpha}$ and solely utilizing the non-attribute latent vector $z^{\beta}$ for classification. The classification performance using the latent vector $\{z^{\beta},z^{\alpha}\}$ is compared, and the results are shown in Table.\ref{tab2:abla}. 

\begin{table}[ht]
  \caption{ Classification ability of different latent vectors \\ on the validation set.}
	\centering
	\begin{tabular}{p{22mm}|lll}
        \hline
		\makecell{\textbf{Models}} & \textbf{Precision} & \textbf{Recall} &\textbf{AUC}\\
        \hline
        \makecell{AGEDR \\ NN(\{$z^{\beta},z^{\alpha}$)\}} & 92.52\%  & \textbf{99.00\%}  & 95.50\%  \\ 
        \hline
        \makecell{AGEDR \\ NN($z^{\beta}$)}  & 91.45\%  & 75.87\%  & 84.02\%  \\ 
        \hline
	\end{tabular}
	\label{tab2:abla}
\end{table}

The precision dropped down to 91.45\%. Notably, a decline in prediction performance is observed, with a particularly significant drop in recall, which decreased to 75.87\%. This suggests two key insights: (1) The AME module carries information related to different attributes that indicate domain shift, and (2) the $L_{disen}$ term successfully disentangles the $z^{\alpha}$ and $z^{\beta}$ components.

\begin{figure}[htb]

\begin{minipage}[b]{0.32\linewidth}
  \centering
  \centerline{\includegraphics[width=2.5cm]{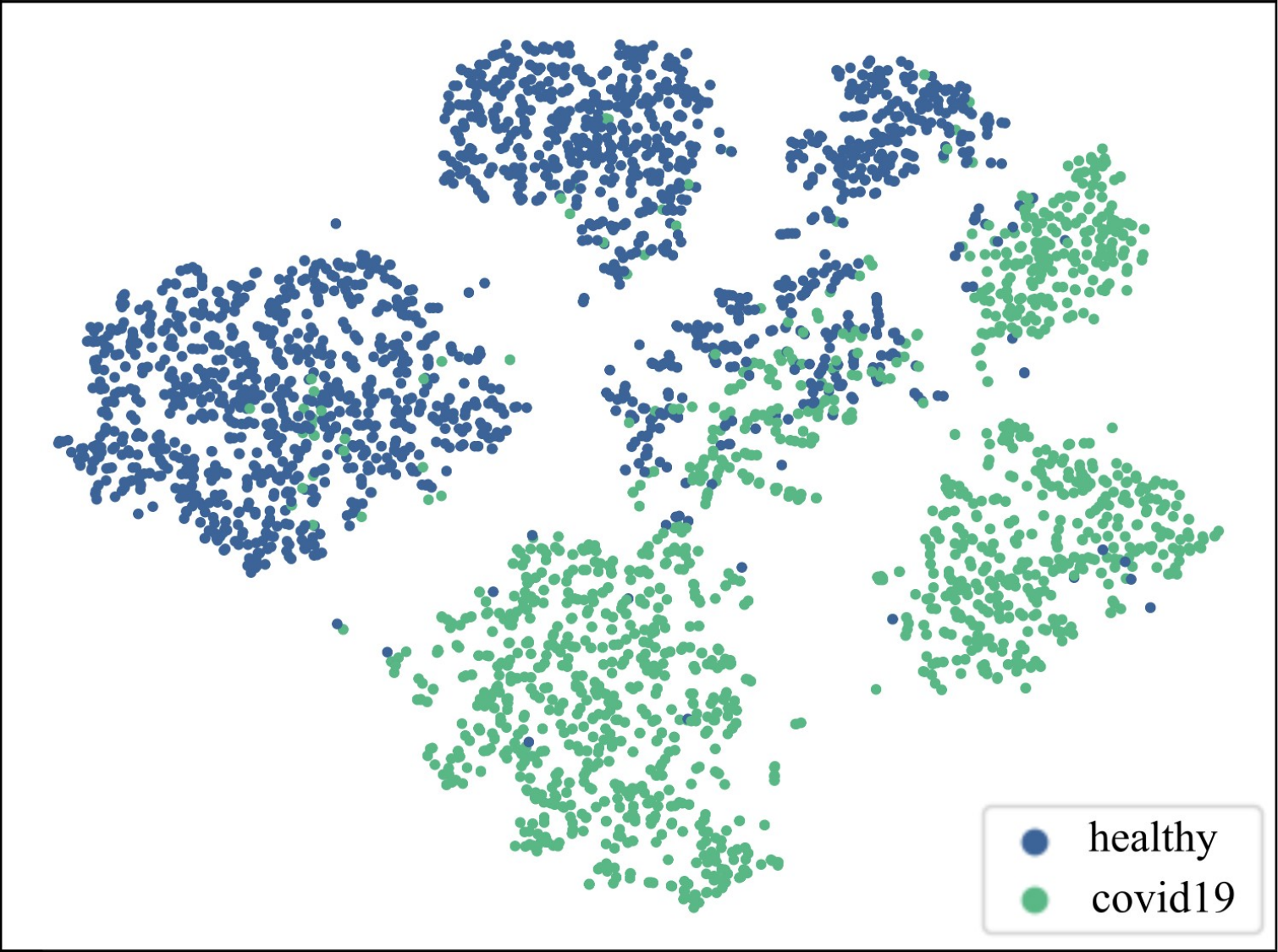}}
  \centerline{(a)}\medskip
\end{minipage}
\begin{minipage}[b]{0.32\linewidth}
  \centering
  \centerline{\includegraphics[width=2.5cm]{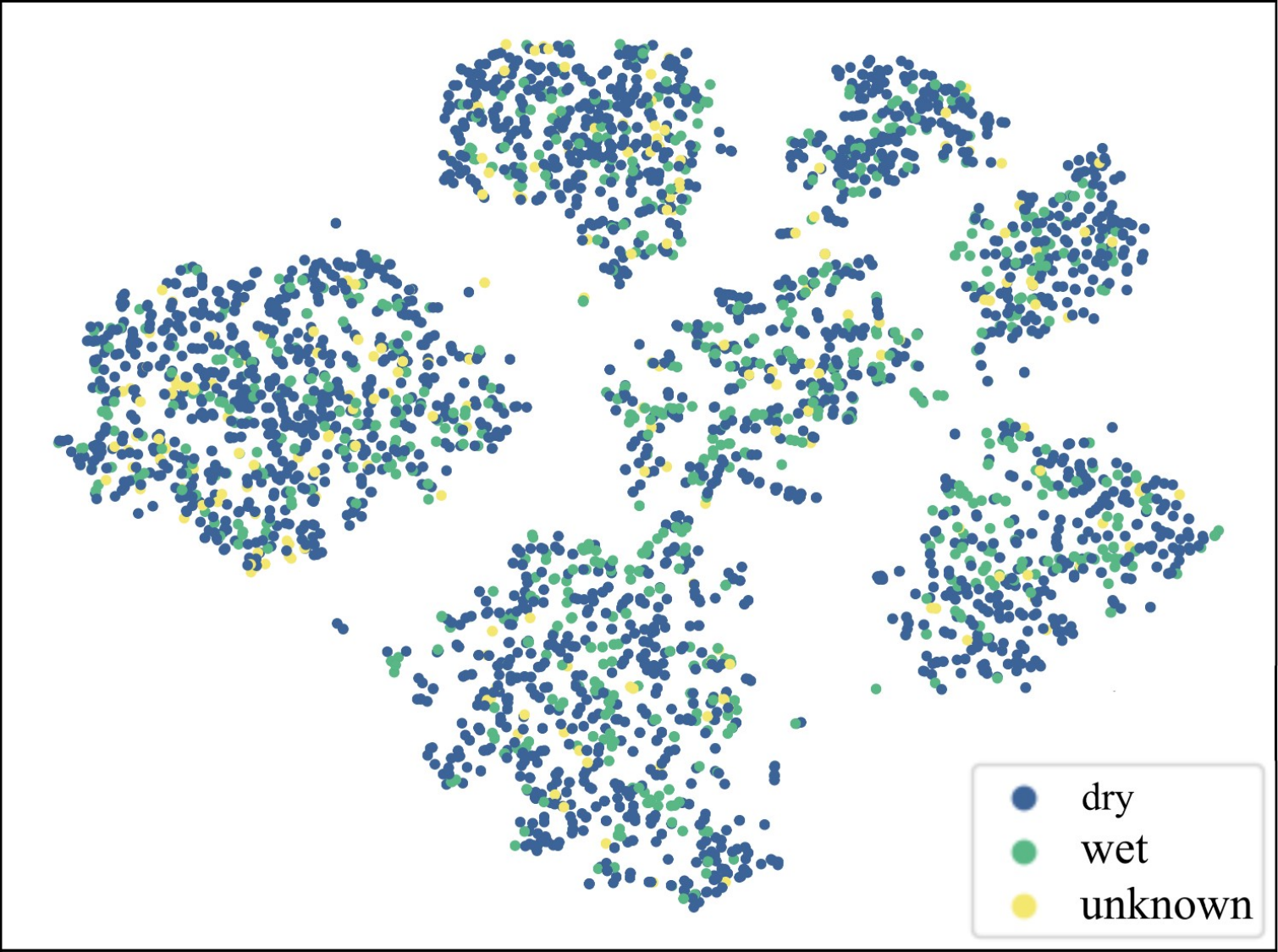}}
  \centerline{(b)}\medskip
\end{minipage}
\hfill
\begin{minipage}[b]{0.32\linewidth}
  \centering
  \centerline{\includegraphics[width=2.5cm]{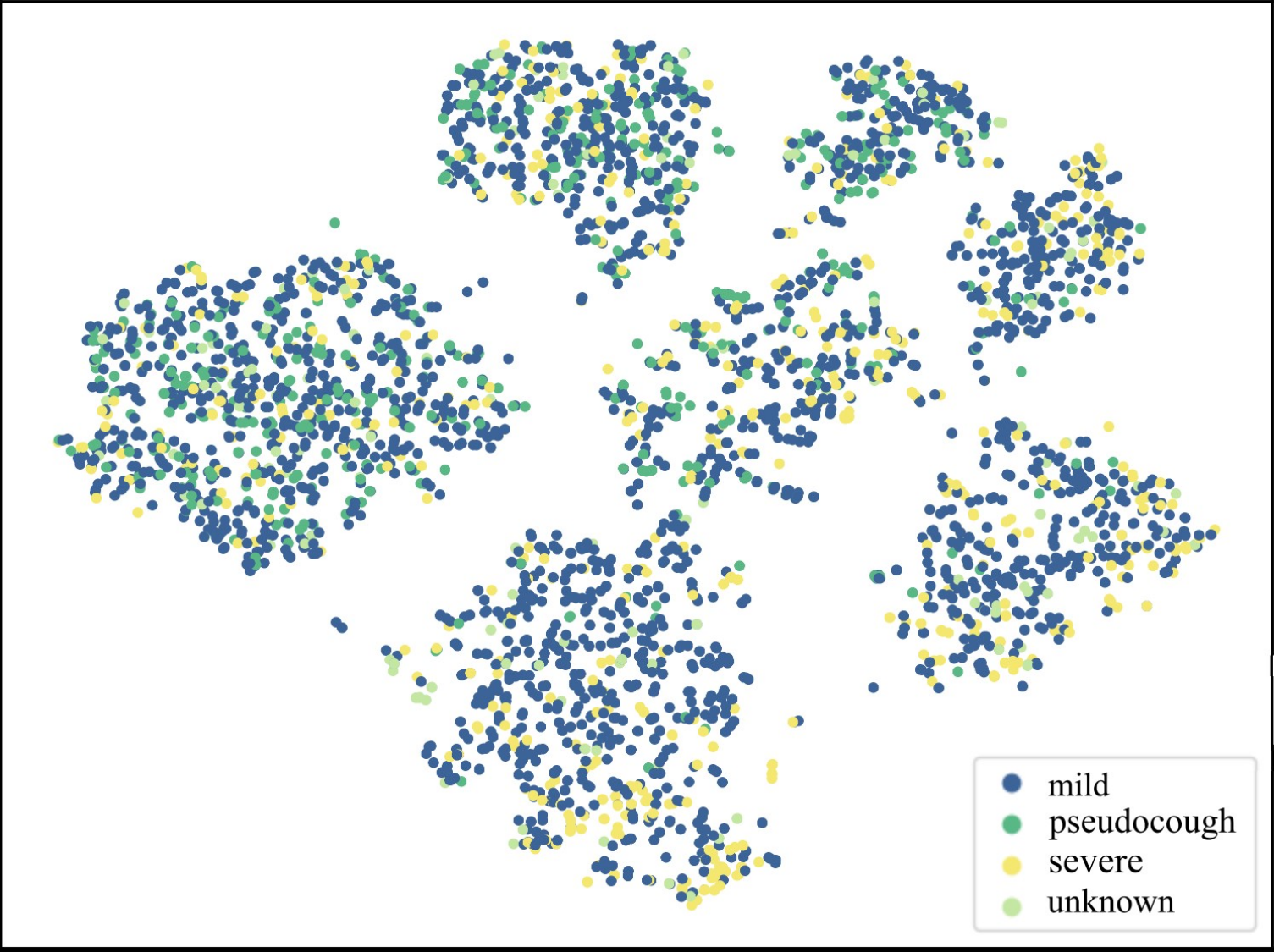}}
  \centerline{(c)}\medskip
\end{minipage}

\caption{The t-SNE visualization of $z=\{z^{\beta}, z^{\alpha}\}$ under labels and different attributes. Where (a) represents the distribution of $z$ under classification labels, (b) denotes the distribution of $z$ under the attribute 'cough\_type', and (c) shows the distribution of $z$ under the attribute 'severity'.}
\label{fig3:disen}
\end{figure}

To validate the fairness of the model, we visualize the latent vectors through t-SNE dimensionality reduction clustering. As shown in Figure.\ref{fig3:disen}, it is evident that the disease classification results correspond solely to the classification labels, whereas the attribute annotations are evenly distributed across the entire sample space, exhibiting no discernible clustering trend. Therefore, AGEDR has been proven to have an attribute desensitization effect on the classification result.

\subsection{Alignment Effect}
In this paper, we use the AME module to map attributes into the implicit vector space to get $\hat (z^{(i)})$, which is converted from scalar to vector through a series of transformations through attribute labels, so it is naturally separable in theory. By using KL divergence to align it with a part of the VAE latent vector, we can make the VAE have the ability to extract the attribute latent vector, and after the adjustment of the loss function, its separability should be better. In order to detect the effect, we use the k-means algorithm to cluster the latent vector $z^{(i)}$. Through experiments, our k-means algorithm achieves 99.21\% clustering accuracy for both cough\_type and severity attributes.

\section{Conclusion}
This paper proposes a disentangled representation learning method named AGEDR that not only disentangles attributes in the latent vectors but also maintains classification accuracy. Specifically, we introduce the AME module, which maps scalar attributes into vectors and aligns them with the VAE latent vectors through an information-theoretic approach. Furthermore, we disentangle the latent vectors of VAE using mutual information. This method holds significant importance in medical audio signal diagnosis, offering interpretability, fairness, and reliability. In the future, we will use the deep Gaussian mixture model to deal with unstructured data and integrate its probability estimation parameters into the hidden vector, so as to mine high-order combination features and achieve stronger disentanglement capability and diagnostic performance on more complex data.

\bibliographystyle{IEEEtran}
\bibliography{IEEEexample}
\end{document}